\documentclass[letterpaper]{article}
\usepackage{aaai}
\usepackage{times}
\usepackage{helvet}
\usepackage{courier}
\usepackage{makeidx}
\usepackage{amssymb,latexsym,amsmath}     
\usepackage{graphicx,array}
\usepackage{enumitem}
\usepackage{booktabs}
\usepackage{multirow} 
\usepackage{multicol} 
\usepackage{float}
\usepackage{algorithm}
\usepackage{algorithmic}
\usepackage{amssymb}
\usepackage{textcomp}
\usepackage[autostyle]{csquotes}
\usepackage{commath}
\usepackage{hhline}
\usepackage{soul}
\usepackage{rotating}
\def\reftable#1{\mbox{Table\,\ref{#1}}}

\newcommand{\napprox}{\not\approx}

\def\shiq{$\mathcal{SHIQ}$}
\def\bfshiq{${\bf\mathcal{SHIQ}}$}
\def\o{$\mathcal{O}$}

\def\L{$\mathcal{D}$}

\def\RL{$\mathcal{RD}$}
\def\IM{\Im}

\begin{document}

\title{ Redundancy-free Verbalization of Individuals for Ontology Validation}

\author{Vinu E.V \and P Sreenivasa Kumar\\
Artificial Intelligence and Databases Lab\\
Department of Computer Science and Engineering\\
Indian Institute of Technology Madras, Chennai, India\\
}
\maketitle

\begin{abstract}
\begin{quote}

We investigate the problem of verbalizing Web Ontology Language (OWL) axioms of domain ontologies in this paper. The existing approaches address the problem of fidelity of verbalized OWL texts to OWL semantics by exploring different ways of expressing the same OWL axiom in various linguistic forms. They also perform grouping and aggregating of the natural language (NL) sentences that are generated corresponding to each OWL statement into a comprehensible structure. However, no efforts have been taken to try out a semantic reduction at logical level to remove redundancies and repetitions, so that the reduced set of axioms can be used for generating a more meaningful and human-understandable (what we call redundancy-free) text. Our experiments show that, formal semantic reduction at logical level is very helpful to generate redundancy-free descriptions of ontology entities. In this paper, we particularly focus on generating descriptions of individuals of \shiq~based ontologies. The details of a case study are provided to support the usefulness of the redundancy-free NL descriptions of individuals, in knowledge validation application. 
\end{quote}
\end{abstract}

\section{Introduction}
\noindent 
Description Logic based ontologies like Web Ontology Language (OWL) ontologies are structures which help in representing the knowledge of a domain, in the form of logical axioms; so that, an intelligent agent with the help of a reasoning system, can make use of them for several applications.  

As an ontology evolves over a period of time, it can grow in size and complexity and unless the updates are carefully carried out, its quality might degrade. 
To prevent such quality depletion, usually an ontology development cycle is accompanied by a validation phase, where both the  knowledge engineers and domain experts meet to review the status of the ontology.

In a typical validation phase, new axioms are included or existing axioms are altered or removed, to maintain the correctness of the ontology. Even though there are automated methods for generating new OWL axioms from a given knowledge source~\cite{pe}, the conventional method for incorporating new axioms and validating the ontology involves a validity check by domain experts. Domain experts, who do the validity check, cannot be expected to be highly knowledgeable  on formal methods and notations. For their convenience, the OWL axioms will have to be first  
converted  into corresponding natural language (NL) texts.  
Ontology verbalizers and ontology authoring tools such as ACE~\cite{kaljurand2007}, NaturalOWL~\cite{Androutso} and SWAT Tools~\cite{third2011}, 
can be utilized for generating controlled natural language (CNL) descriptions of OWL statements. But the verbatim fidelity of such descriptions to the underlying OWL statements, makes them a poor choice for ontology validation. The reason is that, the descriptions  can be confusing to a person who is not familiar with formal constructs, and it is somewhat difficult to correctly make up the meaning from such descriptions. This issue had been previously reported in papers such as~\cite{StevensMWPT11,third2011}, where the authors tried to overcome the issue by applying operations such as grouping and aggregation  on the verbalized text. But, since the fidelity is being treated  at the NL text level, the opportunity for a semantic reduction of the OWL statements to a more meaningful human-understandable representation is abstained.

 For example, consider the following logical axioms (from People \& Pets ontology\footnote{http://www.cs.man.ac.uk/$\sim$horrocks/ISWC2003/Tutorial

/people+pets.owl.rdf}) shown in description logic (DL) and also in the Manchester OWL Syntax\footnote{http://www.w3.org/TR/owl2-manchester-syntax/}. 
\begin{itemize}
\item[1.]\texttt{\small Cat\_Owner $\sqsubseteq$ Person $\sqcap~\exists$hasPet.Animal $\sqcap~\exists$hasPet.Cat}

 \item[2.]\texttt{\small Cat\_Owner(sam)}
\end{itemize}
{\small
\begin{verbatim}
ObjectProperty:<uri#hasPet>
Class: <uri#Person> 
Class: <uri#Animal>
  
Class: <uri#Cat_Owner>
    SubClassOf:
        <uri##hasPet> 
            some <uri#Animal>,
        <uri#Person>
      
Individual: <uri#sam>
     Types: <uri#Cat_Owner>
\end{verbatim}}

  The existing tools generate different variants of the CNL texts like the following as results:
\begin{itemize}
 \item\emph{A cat-owner is a person. A cat-owner has as pet an animal. A cat-owner has as pet a cat. Sam is a cat-owner}. 

or (with grouping and aggregation)

 \item\emph{ A cat-owner is a person . A cat-owner is all of the following: something that has pet an animal, and something that has pet a cat; Examples: sam}.
\end{itemize}
  Even though these texts are not so close to the OWL statements, the usefulness of the description is hindered by the redundancies present in the text. 
  
We propose a system which takes care of the required additional processing of restrictions,  such that the redundant (portion of the) restrictions can be removed, to generate a more semantically comprehensible description. From an application point of view, in this paper, we particularly focus on generating textual description of \emph{individuals} for validating (\shiq~based) ontologies. 

Descriptions of individuals are currently generated by giving importance to technical correctness of the text,   rather than their  naturalness and fluency. For the previous example,  we expect the system to produce a text similar to: \emph{Sam: is a cat-owner having at least one cat as pet}; such that the redundant portion of the text \emph{has as pet an animal} (since it is clear for an expert to imply it from \emph{having at least one cat as pet}) can be removed. 

In the empirical evaluation section, we seek to validate the following two propositions using a case study.  
Firstly, semantic level reduction of redundancies and repetitions can significantly improve the clarity of the domain knowledge descriptions. Secondly, NL descriptions of the individuals of an ontology is useful in validating a given knowledge source. 

\section{Related Work}
Over the last decade, several CNLs such as Attempto Controlled English (ACE)~\cite{kaljurand2007,kaljurand:phd}, Ordnance Survey's Rabbit (Rabbit)~\cite{HartDG07}, and Sydney OWL Syntax (SOS)~\cite{CreganSM07}, have been specifically designed for ontology language OWL. All these languages are meant to make the interactions with formal ontological statements easier and faster for users who are unfamiliar with formal notations. Unlike the other languages~\cite{effectivenl,Jarrar06multilingualverbalization,Androutso} that have been suggested to represent OWL in controlled English, these CNLs are designed to have  formal language semantics and bidirectional mapping between NL fragments and OWL constructs.
Even though these formal language semantics and bidirectional mapping are helpful in enabling a formal check that the resulting NL expressions are unambiguous, they generate a collection of unordered sentences that are difficult to comprehend. 

To use these CNLs as a means for ontology authoring and for knowledge validation purposes, appropriate organization of the verbalized text is necessary. A detailed comparison of the systems that comprehend the NL texts is given in~\cite{StevensMWPT11}. Among such systems, SWAT tools~\cite{third2011} are one of the recent and prominent tools which use standard techniques from computational linguistics to make the verbalized text more readable. They tried to give more clarity to the generated text by grouping, aggregation and elision.  The Semantic Web Authoring (SWAT) tools have given much focus for comprehending the linguistic form of the sentences, rather than handling their logical forms, hence have deficiencies in their NL representations.

 In this paper, we show that, by doing a entailment based reduction at the logical level, and then, by doing a NL mapping and enhancement over the reduced formalisms, a more meaningful human-understandable (what we call redundancy-free) representation can be obtained.

\begin{table*}[!ht]
    \begin{minipage}[!t]{.55\linewidth}
   \centering
   \caption {The syntax and semantics of \shiq~concept types\label{tab:t1}}
    {\small{
    \begin{tabular}{ @{}l@{~~} l l@{}}
    \toprule
    Name & Syntax&Semantics \\ 
   \midrule
   atomic concept  & $A$& $A^{\mathcal{I}}$ (given)  \\
   top concept & $\top$ &$\Delta^{\mathcal{I}}$\\
   bottom concept & $\bot$&$\phi$ \\
   negation & $\neg C$ &$\Delta^{\mathcal{I}}\backslash C^{\mathcal{I}}$\\
   conjunction & $C \sqcap D$&$C^{\mathcal{I}} \cap D^{\mathcal{I}}$ \\
   disjunction & $C \sqcup D$ &$C^{\mathcal{I}} \cup D^{\mathcal{I}}$ \\
   existential restriction & $\exists R.C$ &$\{\,x\in \Delta^{\mathcal{I}}\mid\exists y.\langle x,y\rangle      \in R^{\mathcal{I}}\wedge y \in C^{\mathcal{I}}\,\}$\\
   universal restriction & $\forall R.C$ &$\{\,x\in \Delta^{\mathcal{I}}\mid\forall y.\langle x,y\rangle      \in R^{\mathcal{I}}\Rightarrow y \in C^{\mathcal{I}}\,\}$\\
  min cardinality & $\geq nS.C$ &$\{\,x\in \Delta^{\mathcal{I}}\mid\#\{y\mid\langle x,y \rangle\in R^{\mathcal{I}}\}\ge n\,\}$\\
  max cardinality & $\leq mS.C$ &$\{\,x\in \Delta^{\mathcal{I}}\mid\#\{y\mid\langle x,y \rangle\in R^{\mathcal{I}}\}\le m\,\}$\\
  \bottomrule
  \end{tabular}}}
  \end{minipage} 
  \begin{minipage}[!t]{.45\linewidth}
  \centering
  \caption {The syntax and semantics of  \shiq~ontology axioms}
  \label{tab:t2}
  {\small{
   \begin{tabular}{ @{}l@{~~~}l@{~~}ll@{} }
   \toprule
&Name & Syntax&Semantics \\
\midrule 
  &role hierarchy & $R\sqsubseteq S$&$R^{\mathcal{I}}\subseteq S^{\mathcal{I}}$ \\
 TBox  &role transitivity & Tran($R$) &${R^\mathcal{I}} \circ {R^\mathcal{I}}\subseteq {R^\mathcal{I}} $ \\

 &concept inclusion  & $C \sqsubseteq D$ & $C^{\mathcal{I}} \subseteq D^{\mathcal{I}}$ \\
 &concept equality  & $C \equiv D$ & $C^{\mathcal{I}} = D^{\mathcal{I}}$ \\\midrule
 &concept assertion  & $C(a)$& $a^{\mathcal{I}} \in C^{\mathcal{I}}$ \\
ABox &role assertion  & $R(a,b)$ & $\langle a^{\mathcal{I}}, b^{\mathcal{I}} \rangle \in R^{\mathcal{I}}$ \\
&inequality assertion&$a\napprox b$& $a^{\mathcal{I}}\not = b^{\mathcal{I}} $\\
\bottomrule
\end{tabular}}}
    \end{minipage} 
\end{table*}

\section{Preliminaries}

\subsection{\bfshiq~Ontologies}
The description logic (DL) \shiq~is based on an extension of the well-known logic $\mathcal{ALC}$~\cite{alcx}, with added support for role hierarchies, inverse roles, transitive roles, and qualifying number restrictions~\cite{alc2}.

 We assume $N_{C}$ and $N_{R}$ as countably infinite disjoint sets of \emph{atomic concepts} and \emph{atomic roles} respectively. 
A \shiq~\emph{role} is either $R\in N_{R}$ or an \emph{inverse role} $R^{-}$ with $R \in N_{R}$. To avoid considering roles such as  $(R^{-})^{-}$, we define a function Inv(.) which returns the inverse of a role: Inv($R)=R^{-}$ and Inv($R^{-})=R.$

The set of concepts in \shiq~is recursively defined using the constructors in \reftable{tab:t1}, where $A\in N_{C}, C, D$ are concepts, $R, S$ are roles, and $n,m$ are positive integers. A \shiq~based ontology  --- denoted as a pair $\mathcal{O}=(T,A)$, where $T$ denotes  terminological axioms (also known as TBox)  and $A$ represents assertional axioms (also known as ABox) ---  is a set of axioms of the type specified in \reftable{tab:t2}. A role $R$ in $\mathcal{O}$ is \emph{transitive} if Tran($R$) $\in \mathcal{O}$ or Tran($R^{-}$) $\in \mathcal{O}$. Given an $\mathcal{O}$, let $R_{1}\sqsubseteq_{\mathcal{O}} R_{2}$ be the smallest transitive reflexive relation between roles such that $R_{1}\sqsubseteq R_{2}\in \mathcal{O}$ implies $R_{1}\sqsubseteq_{\mathcal{O}} R_{2}$ and $R^{-}_{1}\sqsubseteq_{\mathcal{O}} R^{-}_{2}$. For a \shiq~ontology $\mathcal{O}$, the role $S$ in every concept of the form $\ge nS.C$ and $\le m S.C$ in $\mathcal{O}$, should be \emph{simple}, that is, $R\sqsubseteq_{\mathcal{O}} S$ holds for no transitive role $R$~\cite{Baader}.

The semantics of \shiq~is defined using \emph{interpretations}. An interpretation is a pair $\mathcal{I}=(\Delta^{\mathcal{I}},.^{\mathcal{I}})$ where $\Delta^{\mathcal{I}}$ is a non-empty set called the \emph{domain} of the interpretation and $.^{\mathcal{I}}$ is the \emph{interpretation function}. The function $.^{\mathcal{I}}$ assigns a set $A^{\mathcal{I}} \subseteq \Delta^{\mathcal{I}}$ to every $A\in N_{C}$, and assigns a relation $r^{\mathcal{I}} \subseteq \Delta^{\mathcal{I}} \times  \Delta^{\mathcal{I}}$ to every $r\in N_{R}$. The interpretation of the inverse role $r^{-}$ is $(r^{-})^{\mathcal{I}}:= \{\langle x,y\rangle\mid\langle y,x\rangle \in r^{\mathcal{I}}\}$. The interpretation is extended to concepts and axioms according to the rightmost column of \reftable{tab:t1} and \reftable{tab:t2} respectively, where 
$\#X$ denotes the cardinality of the set $X$.  

We write $\mathcal{I}\models\alpha$, if the interpretation $\mathcal{I}$ satisfies the axiom $\alpha$ (or $\alpha$ is \emph{true} in $\mathcal{I}$). $\mathcal{I}$ is a \emph{model} of an ontology $\mathcal{O}$ (written $\mathcal{I}\models \mathcal{O}$) if $\mathcal{I}$ satisfies every axiom in $\mathcal{O}$. If we say $\alpha$ is entailed by $\mathcal{O}$, or $\alpha$ is a \emph{logical consequence} of $\mathcal{O}$ (written $\mathcal{O}\models\alpha$), then every model of $\mathcal{O}$ satisfies $\alpha$. A concept $C$ is \emph{subsumed} by $D$ w.r.t. $\mathcal{O}$ if $\mathcal{O} \models C \sqsubseteq D$, and $C$ is \emph{unsatisfiable} w.r.t. $\mathcal{O}$ if $\mathcal{O} \models C \sqsubseteq \bot$. \emph{Classification} is the task of computing all subsumptions $A \sqsubseteq B$ between atomic concepts such that $A,B\in N_{C}$ and $\mathcal{O} \models A \sqsubseteq B$; similarly, \emph{property classification} of \o~is the computation of all subsumptions between  properties $R \sqsubseteq S$ such that $R,S\in N_{R}$ and $\mathcal{O} \models R \sqsubseteq S$. 

\subsection{Semantics of newly introduced DL constructs}


In the description generation section, we introduce two new DL constructs, to represent reduced forms of some of the existing logical forms; this subsection describes their semantics.

For a concept $C$ and a role $R$, the semantics of $\IM\,R.C$ is defined using the interpretation $\mathcal{I}=(\Delta^{\mathcal{I}},.^{\mathcal{I}})$ as
\big\{ $x \in \Delta^{\mathcal{I}}\,|\,\exists y. \langle x,y \rangle \in R^{\mathcal{I}} \land y \in C^{I} \land (\forall z.\langle x,z \rangle \in R^{I}\Rightarrow z \in C^{I})$\,\big\} 

From now on, we address the  restrictions of the form $\IM R.C$ as  \emph{non-vacuous universal restrictions}. By the semantics of $\IM R.C(x)$ and $\exists R.C(x)$, at least one property is guaranteed for the instance $x$, therefore, both these restrictions are addressed as \emph{non-vacuous restrictions} in general.

 The semantics of $\exists ! R.C$ is defined as
\big\{ x $\in \Delta^{\mathcal{I}}\,|\,\exists y. \langle x,y \rangle \in R^{I} \land y \in C^{I} \land 
\big(\exists y_{1}.\langle x,y_{1} \rangle \in R^{I} \land \exists y_{2}.\langle x,y_{2} \rangle \in R^{I}\Rightarrow y_{1}=y_{2}\big) $\,\big\}

\subsection{Running Example}
We use the ontology given in Table~\ref{tbox} (extracted from HarryPotter-book ontology\footnote{https://sites.google.com/site/ontoworks/ontologies})
as the running example throughout this paper. We address this ontology as HP ontology in this paper.

\begin{table}[!th]
\centering\caption {The required portion of HarryPotter-book ontology}
\label{tbox}
\begin{tabular}{@{}l@{}}
  \toprule
{\bf TBox:}\\
 \emph{HogwartsStudent~$\equiv$ Student\ $\sqcap ~\forall$hasPet.Creature}\\
 \emph{$~~~~~~~~~~~~~~~~~~~~~~\sqcap ~\exists$hasPet.Pet~$\sqcap \leq 1$hasPet.Creature}\\[.1cm]

\emph{Pet~$\equiv$ Creature $\sqcap$ $\forall$isPetOf.HogwartsStudent ~$\sqcap$}\\
\emph{$~~~~~~~~~~~~~~~~~~~~~~~~~~~~~~~~~~~~~~~\exists$isPetOf.HogwartsStudent}\\[.1cm]

\emph{HogwartsStudent~$\sqsubseteq$ Student}~~~~~\emph{Cat $\sqsubset$ Pet}\\

\emph{Student~$\sqsubseteq$ Human}~~~~~~~~~~~~~~~~~~~~

\emph{Owl~$\sqsubset$ Pet}\\

\emph{HogwartsStudent~$\sqsubseteq$ Gryffindor $\sqcup$ Slytherin}\\


\emph{Pet~$\sqsubseteq$ Creature}~~~~~~~~~~~~~~~~~~~~~~~~

\emph{Muggle~$\sqsubseteq$ Human}\\
Muggle~$\sqcap~$Wizard~$\sqsubseteq~\bot$~~~~~~~~~~~Pet~$\sqcap~$Student ~$\sqsubseteq~ \bot$\\

Owl~$\sqcap$~Cat~$\sqsubseteq~\bot$~~~~~~~~~~~~~~HalfBlood~$\sqcap$ Muggle $\sqsubseteq$ $\bot$\\[.1cm]

















{\bf ABox:}\\
\emph{HogwartsStudent(harrypotter)}\\
\emph{HogwartsStudent(hermionegranger)}\\
\emph{Muggle(hermionegranger)}\\
\emph{Wizard(harrypotter)}\\
\emph{HalfBlood(harrypotter)}\\
\emph{Gryffindor(harrypotter)}\\
\emph{Gryffindor(hermionegranger)}\\
\emph{$\exists$hasPet.Owl(harrypotter)}\\
\emph{$\exists$hasPet.Cat(hermionegranger)}\\

\emph{Owl(hedwig)}\\
\bottomrule
\end {tabular}
\end {table}

\section{Description Generation}

As we mentioned before, we focus on generating descriptions for each of the individuals in a given \shiq~based ontology. 
  To generate the description of  an individual, we associate with it a set that contains the constraints it satisfies as per the ontology.  We call these sets as the \emph{description-sets} of the individuals.

\label{defnls}The description-set (DS) of an instance $x$ (represented as \L($x$)) in the ontology \o~is defined as follows (where $C$ and $R$ are a concept name and a role name respectively in \o, and $m$ and $n$ are positive integers.)
\begin{equation}
\begin{minipage}{8cm}
$\text{\L}(x)=\big\{C\ |\ \text{\o} \models C(x) \big\} \cup
\big\{ \exists R.C\ |\ \text{\o}\models\ \exists R.C(x) \big\} \cup
\big\{ \forall R.C\ |\ \text{\o}\models\ \forall R.C(x) \big\} \cup
\big\{ \le\!nR.C \ |\ \text{\o}\models\ \le\!nR.C(x) \big\} \cup \big\{ \ge\!mR.C \ |\ \text{\o}\models\ \ge\!mR.C(x) \big\}$
\end{minipage}\nonumber
\end{equation}

In~\cite{EV2015}, the authors have introduced a method for generating  DS of individuals --- they call the sets as \enquote{node-label-sets} --- from a given OWL ontology, using simple SPARQL queries and a reasoner. They were generating the description-sets for a different motive --- generating stems of multiple choice questions.  

A DS may contain redundant and repetitive knowledge, which needs to be removed before verbalizing, to improve the readability as well as the clarity of the content. By redundant knowledge we mean those restrictions which are implied by a strict restriction, or those restrictions which can be combined with other restrictions to form a more human-understandable form. 
For example, consider the DS of the individual \emph{harrypotter}, from our running example. 
\begin{equation}
\begin{minipage}{8cm}
\emph{\L(harrypotter) = \{~HogwartsStudent, Student, Human, Wizard, HalfBlood, Gryffindor, $\exists$hasPet.Pet, $\exists$hasPet.Owl,  $\forall$hasPet.Creature, $\leq$1hasPet.Creature~\}}
\end{minipage}\nonumber
\end{equation}

\begin{figure*}[th!]
  \centering
  \includegraphics[width=1\textwidth]{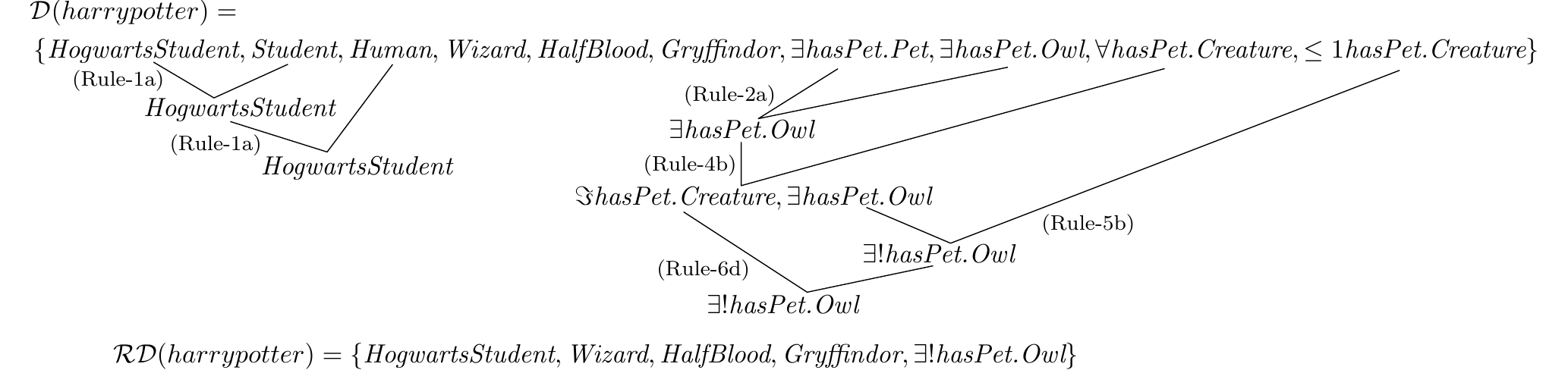}
  \caption{Reduction steps for DS of the individual \emph{harrypotter} from HarryPotter-book ontology \label{abchp}}
\end{figure*}

From the DS, consider the subset \emph{\{~HogwartsStudent, Student, Human \}}. Since \emph{HogwartsStudent $\sqsubseteq$ Student} and \emph{Student $\sqsubseteq$ Human}, the set \emph{\{ Student, Human \}} is a redundant knowledge, and can be removed while generating a description. The reason for considering the set as redundant knowledge is that, being a Hogwarts student clearly implies that Harry Potter is a student and a human. 
Similarly, if $\exists hasPet.Pet$ and $\exists hasPet.Owl$ appear together (given $Owl \sqsubset Pet$), then $\exists hasPet.Pet$ can be be considered as a redundant knowledge.

In order to remove such redundancies and repetitions, we propose  7 sets of entailment based rules that can be applied on the restrictions in a DS. 
The description-sets after applying all the possible rules in the 7 Rule-sets are called Redundancy-free description-set (represented as, $\text{\RL}(x)$).
 
The rules in the 7 Rule-sets are applied in order from Rule-set 1 to  Rule-set 7. 
 The rule-sets and the corresponding rules should be taken in order for reduction; this is because, each rule-set contains carefully chosen restriction patterns whose resulting patterns can be used for further reduction in the forthcoming rule-sets. Moving from a lower rule-set to a higher rule-set, the restrictions which have been applied by a rule can be removed from the DS --- this will greatly reduced the number of combinations of restrictions that are to be considered for applying the rules in the imminent rule-sets.  Considering the size limitation of the paper, we  refrain from explaining the proof of  correctness of the reduction rules here.

\begin{enumerate}
\item {\bf Most-specific concept selection rule}
  \begin{itemize}
  \item[a.] For each class name $V \in$ \L$(x)$, if there exists a $U\in$ \L$(x)$, s.t. \text{\o} $\models  U \sqsubseteq V$,  then add $U$ to \RL$(x)$ and \L$(x)$, and remove $V$ from \RL$(x)$, if present.
  \end{itemize}

\item {\bf Existential class-restrictions' rule}
  \begin{itemize}
  \item[a.] For each $\exists R.U \in$ \L$(x)$, if there exists a $\exists S.V\in$ \L$(x)$, s.t. \text{\o} $\models  U \sqsubseteq V, R \sqsubseteq S$, then add $\exists R.U$ to \RL$(x)$ and \L$(x)$, and remove $\exists S.V$ from \RL$(x)$, if present.
  \end{itemize}

\item {\bf Universal class-restrictions' rules}
  \begin{itemize}
  \item[a.] For each $\forall R.U \in$ \L$(x)$, if there exists a $\forall S.V\in$ \L$(x)$, s.t. \text{\o} $\models  U \sqsubseteq V, S \sqsubseteq R$, then add $\forall R.U$ to \RL$(x)$ and \L$(x)$, and remove $\forall S.V$ from \RL$(x)$, if present.

 \item[b.] For each $\forall R.U \in$ \L$(x)$, if there exists a $\forall S.V\in$ \L$(x)$, s.t. \text{\o} $\models  V \sqsubseteq U, S\equiv R$, then add $\forall R.V$ to \RL$(x)$ and \L$(x)$, and remove $\forall S.V$ from \RL$(x)$, if present.

\item[c.] For each $\forall R.U \in$ \L$(x)$, if there exists a $\forall S.V\in$ \L$(x)$, s.t. \text{\o} $\models  V \sqsubseteq U, R\sqsubset S$, then add $\forall R.V$ and $\forall S.V$  to \RL$(x)$ and \L$(x)$.
  \end{itemize}

\item {\bf I-II Combination rules}
  \begin{itemize}
  \item[a.] For each $\exists R.U \in$ \L$(x)$, if there exists a $\forall S.V\in$ \L$(x)$, s.t. \text{\o} $\models  U \equiv V, S \equiv R$, then add   $\IM R.U$ to \RL$(x)$ and \L$(x)$, and remove $\forall R.U$ and $\forall S.V$ from \RL$(x)$, if present.

 \item[b.] For each $\forall R.U \in$ \L$(x)$,
  if there exists a $\exists S.V\in$ \L$(x)$, s.t. 
  \text{\o} $\models  V \sqsubset U, R\sqsubseteq S$, then add $\IM  R.U$ and $\exists S.V$ to \RL$(x)$ and \L$(x)$, and remove $\forall R.U$ from \RL$(x)$, if present.

\item[c.] For each $\forall R.U \in$ \L$(x)$, if there exists a 
               $\exists S.V\in$ \L$(x)$, s.t. 
     \text{\o} $\models  U \sqsubseteq V, S\sqsubseteq R$, then add $\IM R.U$ and $\IM S.U$  to \RL$(x)$ and \L$(x)$, and remove $\forall R.U$ and $\forall S.V$ from \RL$(x)$, if present.

\item[d.] For each $\forall R.U \in$ \L$(x)$, if there exists a 
               $\exists S.V\in$ \L$(x)$, s.t. 
     \text{\o} $\models  U \sqsubseteq V, R\sqsubset S$, then add $\forall R.U$ and $\exists S.V$  to \RL$(x)$ and \L$(x)$.
  \end{itemize}

\item {\bf Cardinality class-restrictions' rules}
  \begin{itemize}
  \item[a.] For each $\exists R.U \in$ \L$(x)$, if there exists a $\geq$$n S.V\in$ \L$(x)$, s.t. \text{\o} $\models  V \sqsubseteq U, S \sqsubseteq R$ where $n\geq 1$, then add $\geq$$n S.V$ to \RL$(x)$ and \L$(x)$, and remove $\exists R.U$ from \RL$(x)$, if present.

   \item[b.] For each $\exists R.U \in$ \L$(x)$, if there exists a $\leq$$n S.V\in$ \L$(x)$, s.t. \text{\o} $\models  U \sqsubseteq V, R \sqsubseteq S$ where $n=1$, then add $\exists! R.U$ to \RL$(x)$ and \L$(x)$, and remove $\exists R.U$ and $\leq$$nS.V$ from \RL$(x)$, if present.

   \item[c.] For each $\geq$$nR.U \in$ \L$(x)$, if there exists a $\geq$$mS.V\in$ \L$(x)$, s.t. \text{\o} $\models  U \sqsubseteq V, R \sqsubseteq S$ where $n\geq m$, then add $\geq$$mR.U$ to \RL$(x)$ and \L$(x)$, and remove $\geq$$ nS.V$ from \RL$(x)$, if present.
 
  \item[d.] For each $\geq$$nR.U \in$ \L$(x)$, if there exists a $\leq$$nS.V\in$ \L$(x)$, s.t. \text{\o} $\models  U \sqsubseteq V, R \sqsubseteq S$ where $n=1$, then add $\exists! R.U$ to \RL$(x)$ and \L$(x)$, and remove $\geq$$nR.U$ and $\leq$$nS.V$ from \RL$(x)$, if present.
  \end{itemize}

\item {\bf Non-vacuous\footnote{refer the preliminaries section} class-restrictions' rules}
\begin{itemize}
 \item[a.] For each  $\exists R.U \in$ \L$(x)$, if there exists a $\exists! S.V\in$ \L$(x)$, s.t. \text{\o} $\models  U \sqsubseteq V, R \sqsubseteq S$, then add $\exists! R.U$ to \RL$(x)$ and \L$(x)$, and remove $\exists R.U$ from \RL$(x)$, if present.

  \item[b.] For each  $\IM R.U \in$ \L$(x)$, if there exists a $\exists! S.V\in$ \L$(x)$, s.t. \text{\o} $\models  U \sqsubseteq V, R \sqsubseteq S$, then add $\exists! R.U$ to \RL$(x)$ and \L$(x)$, and remove $\IM R.U$ from \RL$(x)$, if present.

    \item[c.] For each  $\exists R.U \in$ \L$(x)$, if there exists a $\exists! S.V\in$ \L$(x)$, s.t. \text{\o} $\models  V \sqsubseteq U, R \sqsubseteq S$, then add $\exists! R.V$ to \RL$(x)$ and \L$(x)$, and remove $\exists R.U$ from \RL$(x)$, if present.

  \item[d.] For each  $\IM R.U \in$ \L$(x)$, if there exists a $\exists! S.V\in$ \L$(x)$, s.t. \text{\o} $\models  V \sqsubseteq U, R \sqsubseteq S$, then add $\exists! R.V$ to \RL$(x)$ and \L$(x)$, and remove $\IM R.U$ from \RL$(x)$, if present.
\end{itemize}

\item {\bf Exactly-one class-restrictions' rules}
\begin{itemize}
 \item[a.] For each  $\exists! R.U \in$ \L$(x)$, if there exists a $\exists! S.V\in$ \L$(x)$, s.t. \text{\o} $\models  U \sqsubseteq V, R \sqsubseteq S$, then add $\exists! R.U$ and $\exists! S.U$ to \RL$(x)$ and \L$(x)$, and remove $\exists !S.V$ from \RL$(x)$, if present.

\item[b.] For each  $\exists! R.U \in$ \L$(x)$, if there exists a $\exists! S.V\in$ \L$(x)$, s.t. \text{\o} $\models  V \sqsubseteq U, R \sqsubseteq S$, then add $\exists! R.V$ and $\exists! S.V$ to \RL$(x)$ and \L$(x)$, and remove $\exists R.U$ from \RL$(x)$, if present.

\end{itemize}

\end{enumerate}
Figure~\ref{abchp} shows the reduction steps of \L(\emph{harrypotter}). The constraints in the DS are taken two at a time and  we consider the possible applications of the rules in the Rule-sets 1 to 7. At first, the rule in Rule-set 1 (denoted as, Rule-1a) is  applied repeatedly to obtain the most specific class name. Then, the rule in Rule-set 2 (Rule-2a), the second rule in the Rule-set 4 (Rule-4b), the second rule in Rule-set 5 (Rule-5b), and finally the fourth rule in Rule-set 6 (Rule-6d), are applied in order to reduce the property related restrictions in the DS. 



\section{Linguistic Description of Individuals}
For  the completeness of the paper, we present a simple method which we adopted to generate linguistic descriptions of individuals from their redundancy-free DS. 

 Linguistic description of an individual is defined as the set of NL fragments which describes the class names and property related constraints it satisfies. An example of a description of Harry Potter  (individual \emph{harrypotter}) from HP ontology 
is given as: 
\begin{equation}
\framebox{
\begin{minipage}{8cm}
\texttt{Harry Potter:} { is a Hogwarts Student, a Wizard,
a Halfblood, a Gryffindor and having exactly one Owl as Pet}
\end{minipage}
}\nonumber
\end{equation}

We consider a template similar to the following regular expression (abbreviated as regex) for generating an individual's description.
\begin{equation}
\begin{minipage}{8cm}
\texttt{Individual:} (\enquote{is}) \big((\enquote{a}) ClassName (\enquote{,} $|$ \enquote{and})$?\big)^{+}$ \big((PropertyRestriction)(\enquote{,} $|$ \enquote{and})$?\big)^{+}$
\end{minipage}\nonumber
\end{equation}

\begin{table*}[!th]\centering\caption{Examples of the descriptions of individuals that are generated using our proposed approach and the traditional approach from  PP, HP and GEO ontologies.\label{eg1}}
\begin{tabular}{@{}p{7.5 cm}p{8 cm}p{1.5cm}@{}}\toprule
\emph{Proposed approach}&\emph{Traditional approach}&\emph{Ontology}\\\midrule

\emph{Bird cherry Oat Aphid:} is an insect damage, having  at least one pest  and only pest as factor.& \emph{Bird cherry Oat Aphid:} is a disorder, bio-disorder, pest damage and insect damage. It is all the following: has as factor only pest-insect, has as factor only pest, has as factor only organism and has as factor something.&PP\\\midrule

\emph{Black Chaff:} is a plant bacterioses, having at least one microorganism and only microorganism as factor. & \emph{Black Chaff:} is a disorder, a biotic disorder and a plant bacterioses. It is all the following: has as factor bacterioses, has as factor only organism, has as factor at least 1 thing, has as factor only micro-organism.&PP \\\midrule

\emph{Hermione Granger:} is a Hogwarts Student, a muggle, a gryffindor, having exactly one cat as pet.

&\emph{Hermione Granger:} is a Hogwarts student, a student, a human, a muggle, a gryffindor. It is all the following: has as pet a pet, has as pet a cat, has as pet only creature, has at least 1 creature, as pet.  &HP  \\\midrule

\emph{Hedwig:} is an owl, is related to at least one Hogwarts student and only Hogwarts student, as pet.

&\emph{Hedwig:} is an owl, a pet, a creature. It is all the following: is pet of only Hogwarts student, is pet of a Hogwarts student.&HP \\\midrule

\emph{Jersey:} is a geopolitical dependency and is related to exactly one sovereign state as a member.
&\emph{Jersey:} is a geopolitical dependency, an organization, a governmental organization, an Independent continuant, a subnational entity  and is a member of exactly one sovereign state.
&GEO\\\midrule
\emph{Florida:} is a major administrative subdivision, is related to at least one nation as a part, and is related to exactly one sovereign state as a member.
&\emph{Florida:} is a major administrative subdivision, an organization, a governmental organization, an Independent continuant, a subnational entity. It is all the following: is a part of at least one nation, and is a member of exactly one sovereign state.
&GEO\\

\bottomrule
\end{tabular}
\end{table*}

In the above regex, ClassName specifies the concept names in the DS. We use the \texttt{rdfs:label} property values of the class names as the ClassName. If \texttt{rdfs:label} property is not available, the local names of the URIs are used as the ClassName. For PropertyRestriction, the property related class restrictions in the DS are utilized. The property related constraints are treated in parts. We first tokenize the property names in the constraints. Tokenizing includes word-segmentation and processing of camel-case, underscores, spaces, punctuations etc. Then, we identify and tag the verbs\footnote{In the absence of a proper verb, the phrase \enquote{related to} is used in its place.} and nouns in the segmented phase --- as R-\texttt{verb}, R-\texttt{noun} respectively --- using the Natural Language Tool Kit\footnote{Python NLTK: http://www.nltk.org/}.  Some of these R-\texttt{verb} are given pre-defined morphological word forms. For example, the verb \lq has\rq~will be changed to \lq having\rq. We then incorporate these segmented words in a \emph{constraint-specific template}, to form a PropertyRestriction. For instance, the restriction \emph{$\exists$hasPet.Cat} is verbalized to \enquote{having at least one pet as cat}, using the template: {$<\!\text{R-}\texttt{verb}\!>$ at least one $<\!C\!>$ as $<\!\text{R-}\texttt{noun}\!>$}. Constraint-specific templates corresponding to the possible restrictions in a DS are listed in Table-\ref{t1}. Linguistic variations of these constraint-specific templates are also possible, to enhance the readability. But, since the empirical study (see the next section) is done for a different intention with the help of a carefully chosen participants, we limit further fluency enhancement of the texts.

\begin{table}[!ht]\caption{Constraint-specific templates of the possible restrictions in a redundancy-free description-set.\label{t1}}
\centering
\begin{tabular}{@{}l@{ }l@{}}\midrule
Restrictn. & Constraint-specific template\\\midrule
$\exists R.C$ & $<\!\text{R-}\texttt{verb}\!>$ at least one $<\!C\!>$ as $<\!\text{R-}\texttt{noun}\!>$\\
$\forall R.C$ & $<\!\text{R-}\texttt{verb}\!>$ only $<\!C\!>$ as $<\text{R-}\texttt{noun}>$\\
$\ge\!nR.C$ & $<\!\text{R-}\texttt{verb}\!>$ at least $<\!n\!><\!C\!>$ as $<\!\text{R-}\texttt{noun}\!>$\\
$\le\!mR.C$ & $<\!\text{R-}\texttt{verb}\!>$ at most $<\!m\!><\!C\!>$ as $<\!\text{R-}\texttt{noun}\!>$\\
$\IM R.C$ &  $<\!\text{R-}\texttt{verb}\!>$ at least one $<\!C\!>$ and\\&~~~~~~~~~~~~~~~~~~~~ only $<\!C\!>$ as $<\!\text{R-}\texttt{noun}\!>$\\

$\exists !R.C$ & $<\!\text{R-}\texttt{verb}\!>$ exactly one $<\!C\!>$ as $<\!\text{R-}\texttt{noun}\!>$\\\midrule
\end{tabular}
\end{table}

We avoid the restrictions that contain $\top$ (apex class) or $\bot$ (bottom class), for generating the description; this is purely a design decision. The inclusion of such restrictions may force to consider new cases of the constraint-specific templates, in addition to what is given in Table-\ref{t1}, which we are not right now interested in.

\section{Empirical Evaluation}
We present a case study to explore the applicability of the redundancy-free description of instances in validating the domain knowledge. Rather than choosing an ontology under development, we study the case of validating a previously built ontology. 
Plant-protection ontology\footnote{https://sites.google.com/site/ppontology/} (a.k.a. PP ontology), an ontology which had been used for the empirical study in~\cite{EV2015}, is chosen for our study. 

In the study, domain experts were presented with two representations of the same knowledge: one is by direct verbalization of the description-sets  and the other is by verbalizing them after finding the corresponding redundancy-free description-sets. Direct verbalization of a DS generates texts (or descriptions) which are similar to those texts  which are produced by an existing ontology verbalizer --- we call this method as \emph{traditional approach}, and the other as the \emph{ proposed approach}. Examples for  the description texts that are generated using the proposed approach and traditional approach, from the PP ontology, HP ontology and Geographical Entity\footnote{https://bitbucket.org/uamsdbmi/geographical-entity-ontology/src (last accessed: 27/11/2015)} 
(GEO) ontologies are 
 given in Table~\ref{eg1}.

The experts were then asked to mark their degree of understanding of the knowledge in the scale: (1) poor; (2) medium; (3) Good. 

To measure the usefulness of our approach in validating the domain knowledge; corresponding to each of the instance-descriptions, domain experts are told to choose from the options: (1) Valid (2) Invalid (3) Don’t know (4) Cannot be determined. Also, feedback is being taken to get suggestions on improving the system.

  PP-ontology has 546 instances, 105 concepts and 15 object properties. Corresponding to each of the 546 instances, we have generated 546 description texts, using an implemented prototype of the system. Since manual evaluation of all the generated descriptions is difficult, we grouped the instances based on their (redundancy-free) description-set, and aggregated the names of the instances using suitable conjunctions (for e.g., Yellow rust \emph{and} Brown rust). An ontology description, containing 31 sentences, has been obtained for evaluation. Three experts on plant protection related area reviewed the verbalized descriptions.

\subsection{Results}

\subsubsection{Does it improve the understandability?} Degree of understanding of each of these descriptions to a domain expert, is identified by looking at the options (poor, medium or good) which she had chosen. If there exists an ambiguity in the description (due to its verbatim fidelity to OWL statements), she is expected to choose poor or medium level as the understanding. To confine the reasons for ambiguity to the fidelity  to OWL alone, possible (manual) editing had been done on the generated text --- as we are not using any sophisticated NL generation techniques.

Table~\ref{response1} shows the statistics of the responses which we received for the descriptions that are generated using our approach, from three domain experts. Overall response (fourth row) in the table corresponds to the response of the majority (at least 2 out of 3). For those descriptions which are generated from redundancy-free DS,  24 out of 31 texts are rated as \enquote{good}, whereas for those which are generated directly from DS, only 5 out of 31 texts are rated as \enquote{good} (see Table~\ref{response2}). This highlights the significance of the redundancy reduction process, in domain knowledge understanding. 

\subsubsection{How helpful is it in knowledge validation?} 

Usefulness of the generated descriptions in validating an ontology, can be obtained by looking at the number of description texts which are marked as \enquote{Cannot be determined}. The three options: Valid, Invalid and Don't know, imply that the text is useful in getting into a conclusion, whereas the option \enquote{Cannot be determined} indicates some problem in the representation. In Table~\ref{response3}, we show the count of the description texts which have been marked as \enquote{Cannot be determined} by the majority of the domain experts, for the traditional and proposed approaches. In case of the proposed approach, only  2 out of 31 descriptions are not useful in determining the quality of the ontology, whereas in case of the traditional approach, approximately 50 percentage of the descriptions are not helpful. 

\subsubsection{Domain Experts feedback and discussion} 

The participants agree with the fact that, by reducing the redundancies in a description, the amount of time required for validating an instance description is reduced to a great extent.

Validation of an ontology also involves verifying the truthfulness of the property relationships in it, which is not addressed in this paper. This issue can be addressed in future by generating description-sets for pairs of instances, and mapping them to the respective constraint(s) in the DS of the first instance. For e.g., $\text{\L}(a)=\{ C_{1},C_{2},\exists hasFriend.C_{3}\}$, and $\text{\L}(a,b)=\{R\}$, then $hasFriend$ in $\text{\L}(a,b)$ can be mapped to $\exists hasFriend.C_{3}$ in $\text{\L}(a)$. The description of $a$ can be generated as \enquote{$a$: is a $C_{1}$ and $C_{2}$, and having some $C_{3}$, like $b$, as Friend.} 

According to the domain experts, a persisting problem with any validation phase (especially when it involves  instance-wise description generation and experts validating the verbalized knowledge) is that, when the ontology  becomes very large and complex, validation phase becomes a bottleneck for the entire development cycle. One way to overcome this issue in our validation approach is by considering only a relevant subset of instances and their descriptions, so that, a rough estimate of the erroneous formalisms in the ontology can be identified instantly. 



\begin{table}[!th]\caption{Statistics of the responses which we received for the 31 text descriptions that are generated using our \emph{ proposed approach}.\label{response1}}\centering
\begin{tabular}{@{}crr@{}r@{}}\toprule
\multirow{2}{*}{Domain Expert No.}&\multicolumn{3}{c}{No. of sentences with rating:} \\ 
&\emph{Poor}& \emph{~~~Medium}& \emph{Good}\\\midrule
1&4&4&23\\
2&6&3&22\\
3&2&4&25\\\midrule
Majority Responses:&5&2&24\\
\bottomrule
\end{tabular}
\end{table}

\begin{table}[!th]\caption{Statistics of the responses which we received for the 31 text descriptions that are generated using the \emph{traditional approach}.\label{response2}}\centering
\begin{tabular}{@{}crrr@{}}\toprule
\multirow{2}{*}{Domain Expert No.}&\multicolumn{3}{c}{No. of sentences with rating:} \\ 
&\emph{Poor}& \emph{~~~Medium}& \emph{Good}\\\midrule
1&11&13&7\\
2&8&18&5\\
3&12&11&8\\\midrule
Majority Responses:&13&13&5\\
\bottomrule
\end{tabular}
\end{table}

\begin{table}[!th]\caption{Statistics (based on the majority responses) to determine the usefulness of the generated text descriptions in validating the ontology.\label{response3}}\centering
\begin{tabular}{@{}lrrrr@{}}\toprule
\multirow{3}{*}{Approach}&\multicolumn{4}{c}{No. of descriptions that are marked as:}\\
&\multirow{2}{*}{\emph{Valid}} &\multirow{2}{*}{\emph{~Invalid}}&\emph{~Don't}&\emph{Cannot be}\\
&      &       &\emph{~know}&\emph{determined}\\\midrule
Proposed   &26 &2 &0 &2\\
Traditional&12 &1 &3 &15\\
\bottomrule
\end{tabular}
\end{table}

\section{Conclusion}
A novel method for generating   text descriptions of individuals of a given \shiq~ontology is proposed in the paper. The descriptions are not verbatim translations of logical axioms of the ontology. Rather, they are generated from a description of the individual on which semantic simplification has been carried out. We propose entailment-based reduction rules for this purpose. We find that the proposed method indeed gives redundancy-free descriptions of individuals.

Empirical studies based on a rather small ontology 
 show that the redundancy-free description of the domain knowledge is helpful in understanding the formalized knowledge more effectively and also useful in validating them.

As a future work, we plan to implement a \emph{Protege} plug-in to allow ontology developers to benefit from the suggested approach.




\selectfont\fontsize{9.5pt}{10.5pt}
\bibliography{ref.bib}{}
\bibliographystyle{aaai}

\end{document}